\documentclass[conference]{IEEEtran}
\IEEEoverridecommandlockouts
\usepackage{cite}
\usepackage{amsmath,amssymb,amsfonts}
\usepackage{graphicx}
\usepackage{textcomp}
\usepackage{xcolor}
\def\BibTeX{{\rm B\kern-.05em{\sc i\kern-.025em b}\kern-.08em
    T\kern-.1667em\lower.7ex\hbox{E}\kern-.125emX}}

\usepackage{hyperref}
\usepackage{amsthm}

\usepackage{algorithm}
\usepackage{algpseudocode}
\usepackage{tabularx}
\usepackage{enumitem}
\usepackage{multirow}
\usepackage{wrapfig}
\usepackage{bm}
\usepackage{booktabs}

\usepackage{siunitx}
\newcolumntype{d}{S[table-format=3.2(3), separate-uncertainty]}

\newtheorem{definition}{Definition}

\makeatletter
\def\@opargbegintheorem#1#2#3{\trivlist
   \item[]{\bfseries #1\ #2\ (#3)} \itshape}
\makeatother

\makeatletter
\newcommand{\multiline}[1]{%
  \begin{tabularx}{\dimexpr\linewidth-\ALG@thistlm}[t]{@{}X@{}}
    #1
  \end{tabularx}
}
\makeatother

\begin{document}

\title{Feature-Space Semantic Invariance: Enhanced OOD Detection for Open-Set Domain Generalization}

\author{
\IEEEauthorblockN{Haoliang Wang$^1$, Chen Zhao$^2$, Feng Chen$^1$}
\IEEEauthorblockA{
$^1$\textit{Department of Computer Science, The University of Texas at Dallas, Richardson, Texas, USA} \\
$^2$\textit{Department of Computer Science, Baylor University, Waco, Texas, USA}\\
\{haoliang.wang, feng.chen\}@utdallas.edu, chen\_zhao@baylor.edu}
}

\maketitle

\begin{abstract}
    Open-set domain generalization addresses a real-world challenge: training a model to generalize across unseen domains (domain generalization) while also detecting samples from unknown classes not encountered during training (open-set recognition). However, most existing approaches tackle these issues separately, limiting their practical applicability. To overcome this limitation, we propose a unified framework for open-set domain generalization by introducing \textit{Feature-space Semantic Invariance} (FSI). FSI maintains semantic consistency across different domains within the feature space, enabling more accurate detection of OOD instances in unseen domains. Additionally, we adapt a generative model to produce synthetic data with novel domain styles or class labels, enhancing model robustness. Initial experiments show that our method improves AUROC by 9.1\% to 18.9\% on ColoredMNIST, while also significantly increasing in-distribution classification accuracy.

\end{abstract}


\section{Introduction}

Open-set domain generalization addresses the dual challenge of domain generalization (DG) and open-set recognition (OSR) by aiming to classify in-distribution (ID) instances under domain shifts while simultaneously detecting samples from unknown classes in the unseen target domain. Despite this setting better reflecting real-world scenarios, current methods often treat it as two separate problems due to its complexities. For instance, OOD samples with styles similar to IDs are difficult to detect, while ID samples in a shifted target domain are prone to being misclassified as OOD.

Research in open-set domain generalization is relatively sparse, with only a few works \cite{shao2024supervised,zhao2023open,zhao2024algorithmic,zhao2023towards,he2024gdda,zhao2024dynamic,wang2023towards,wang2024madod,jiang2024feed,wang2022layer} making progress compared to conventional approaches that handle either DG or OSR in isolation. However, these existing methods typically rely heavily on OOD training data \cite{katsumata2021open}, which is rarely available in training, or utilize meta-learning strategies and complex network architectures \cite{shu2021open, wang2023generalizable, noguchi2023simple} that lack compatibility with state-of-the-art OOD detection techniques, such as Energy \cite{liu2020energy} and DDU \cite{mukhoti2023deep}.

To address these limitations, we propose an advanced technique that enforces feature-space semantic invariance to learn high-quality domain-invariant features, and leverages synthetic OOD data to increase the separability between ID and OOD. Our main contributions are as follows:

\begin{itemize}[leftmargin=*]
    \item We introduce Feature-space Semantic Invariance (FSI) to enforce semantic consistency across domains within the feature space. By aligning semantic features across augmented samples, FSI enables the model to learn high-quality domain-invariant features, enhancing its generalizability to unseen domains. 
    \item We incorporate synthetic OODs generated from ID samples to establish clearer decision boundaries between ID and OOD instances, significantly enhancing the model's robustness against OOD.
    \item Preliminary experiments show our approach improves AUROC by 9.1\% to 18.9\% on the ColoredMNIST dataset, with a notable increase in ID classification accuracy. These results validate the model’s potential in open-set domain generalization, positioning it as a viable solution for applications with novel domains and classes.
\end{itemize}

\begin{figure}[!t]
    \centering
    \includegraphics[width=1.\linewidth]{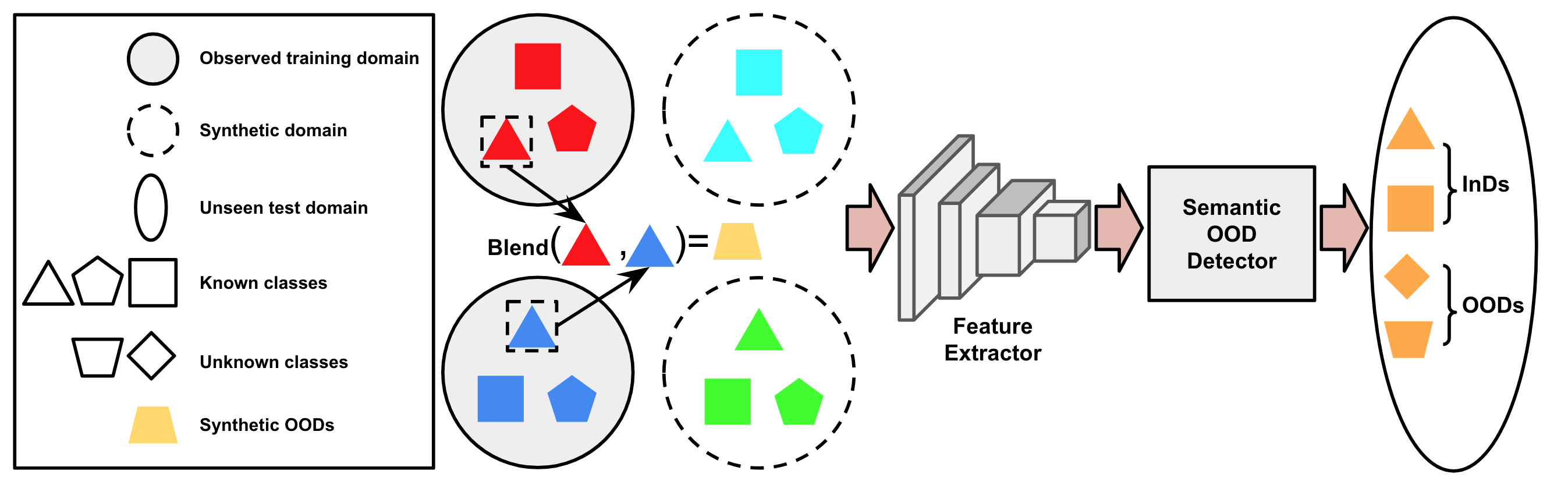}
    \vspace{-6mm}
    \caption{An illustration of the proposed framework. The framework samples instances from different domains, each characterized by unique variations (e.g., colors), aiming to learn a domain-invariant feature extractor that can be combined with state-of-the-art semantic OOD detectors to effectively address both domain generalization and open-set recognition challenges.}
    \label{fig:problem}
    \vspace{-2mm}
\end{figure}
    


\section{Methodology} 

\textbf{Problem Setting.} Given a dataset $\mathcal{D}$, we consider a set of domains $\mathcal{E}=\{e\}_{e=1}^E\in\mathbb{N}$, where each domain corresponds to a data subset $\mathcal{D}^{e}=\{(\mathbf{x}^{e}_i,y^{e}_i)\}_{i=1}^{|\mathcal{D}^{e}|}$ with specific data variations. The dataset $\mathcal{D}$ is thus represented as $\mathcal{D}=\{\mathcal{D}^{e}\}_{e=1}^E$. 
We divide $\mathcal{D}$ into training domains $\mathcal{E}_{tr}\subset\mathcal{E}$ and test domains $\mathcal{E}_{te}=\mathcal{E}\backslash\mathcal{E}_{tr}$. The training data $\mathcal{D}_{tr}$ consists of samples $\mathcal{D}_{tr}=\{\mathcal{D}^e\}_{e=1}^{\mathcal{E}_{tr}}\in(\mathcal{X}\times\mathcal{Y}_{tr})$, where $\mathcal{X}\in\mathbb{R}^d$ represents the input space and $\mathcal{Y}_{tr}=\{1,\cdots,K\}$ denotes the label space, with $K$ being the number of observed classes in $\mathcal{D}_{tr}$. The test domains, $\mathcal{E}_{te}$, are associated with data $\mathcal{D}_{te}=\mathcal{D}\backslash\mathcal{D}_{tr}\in(\mathcal{X}\times\mathcal{Y}_{te})$, where $\mathcal{Y}_{te}=\{\mathcal{Y}_{tr}, OOD\}$, with an unknown class (OOD) label $OOD$.

The main challenge in open-set domain generalization is to develop a network \( f \) that can identify novel classes in shifted domains using a semantic OOD detector \( \omega \). Given that the test data from \( \mathcal{E}_{te} \) is unavailable and novel classes appear only in \( \mathcal{D}_{te} \), this problem is inherently difficult.
We assume that inter-domain variation arises solely from covariate shift \cite{robey2021model}, meaning that domain differences stem from variations in the marginal distributions \( \{\mathbb{P}(X^e)\}_{e \in \mathcal{E}} \), which can be connected by a generative model \( G: \mathcal{X} \times \mathcal{E} \rightarrow \mathcal{X} \), that $\mathbf{x}^{e'}=G(\mathbf{x}^{e},e')$ for all $e,e'\in\mathcal{E}$.

Let the network be defined as \( f := g \circ h \), where \( g: \mathcal{X} \times \Theta \rightarrow \mathbb{R}^r \) is the feature extractor and \( h: \mathbb{R}^r \times \Theta \rightarrow \mathbb{R}^K \) is the classifier. To tackle the open-set domain generalization problem, we enforce the feature-space semantic invariance: the instance-conditional distributions of features extracted by \( g \) should remain stable across different domains. Formally, we define the feature-space semantic invariance (FSI) as follows:

\begin{definition}[Feature-space Semantic Invariance (FSI)]
\label{def:feature_space_inv}
    Given $G$, a network $f=g\circ h$ is said to exhibit feature-space semantic invariance if $g(\mathbf{x}^{e},\boldsymbol{\theta}_g)=g(\mathbf{x}^{e'},\boldsymbol{\theta}_g)$ almost surely, where $\mathbf{x}^{e'}=G(\mathbf{x}^{e},e'),\mathbf{x}^{e}\sim\mathbb{P}(X^{e}), \mathbf{x}^{e'}\sim\mathbb{P}(X^{e'})$, for all $e,e'\in\mathcal{E}$.
\end{definition}


\textbf{Design of Generative Model $G$.} Following the data disentanglement principles of \cite{robey2021model, vapnik1999nature}, which separate domain variations into latent vectors, we consider an ID sample \((\mathbf{x}^e, y^e)\) from a domain \(e \in \mathcal{E}\). It is assumed that the sample is generated from two vectors: a latent semantic vector \(\mathbf{s}\) and a domain-specific latent variation vector \(\mathbf{v}^e\). The generative model \(G\) consists of a semantic encoder \(E_{\text{sem}}: \mathcal{X} \rightarrow \mathcal{S}\), a variation encoder \(E_{\text{var}}: \mathcal{X} \rightarrow \mathcal{V}\), and a decoder \(D: \mathcal{S} \times \mathcal{V} \rightarrow \mathcal{X}\). The model disentangles the input data into semantic and variation vectors, where \(\mathbf{s} = E_{\text{sem}}(\mathbf{x}^e)\) and \(\mathbf{v}^e = E_{\text{var}}(\mathbf{x}^e)\), and reconstructs the input via \(\mathbf{x}^e=D(\mathbf{s}, \mathbf{v}^e)\). Furthermore, \(G\) can generate a synthetic instance $\mathbf{x}^{e'}=D(\mathbf{s},\mathbf{v}^{e'})$ for a domain \(e'\) by substituting \(\mathbf{v}^e\) with a randomly sampled \(\mathbf{v}^{e'} \sim \mathcal{N}(0, \mathbf{I})\).

In this work, \(G\) is treated as a pre-trained model that follows the architecture described in \cite{robey2021model}.


\textbf{Feature Regularization $R_{F}$.} To enforce feature-space semantic invariance (FSI), we propose the following regularization term:
\begin{equation}
\small
\begin{aligned}
\label{eq:reg_dg}
    R_{F} = \mathbb{E}_{(\mathbf{x}^e, y)\sim\mathcal{D}_{tr}} d\Big[g(\mathbf{x}^e, \boldsymbol{\theta}_g), g\big(G(\mathbf{x}^e, \mathbf{v}^{e'}), \boldsymbol{\theta}_g\big) \Big]
\end{aligned}
\end{equation}
where $\mathbf{v}^{e'}\sim\mathcal{N}(0,\mathbf{I})$. We use the $\ell_1$-norm for the distance function $d:\mathbb{R}^r\times\mathbb{R}^r\rightarrow\mathbb{R}$ in $R_{F}$. The regularizer \( R_{F} \) enables \( g \) to capture semantic features by minimizing the distance between outputs of \( g \) for original data and their augmented counterparts in the random domain \(\mathbf{v}^{e'}\), while also mitigating domain shift disruptions during OOD detection by encouraging \( g \) to learn domain-invariant semantic features.

\textbf{Energy Regularization $R_{E}$.}  
For enhanced OOD detection capability within the context of open-set recognition, we generate synthetic OODs using semantic inter-class blending. Given two instances with different class labels \((\mathbf{x}^{e_1}, y_1)\) and \((\mathbf{x}^{e_2}, y_2)\), we compute the augmented semantic representation \(\tilde{\mathbf{s}}\) as follows:
\begin{equation}
\small
\tilde{\mathbf{s}} = \alpha \cdot \mathbf{s}_1 + \beta \cdot \mathbf{s}_2
\label{eq:blending}
\end{equation}

where \(\alpha, \beta \in [-100, 100]\), and \(\mathbf{s}_1, \mathbf{s}_2\) are the semantic vectors of $\mathbf{x}^{e_1}$ and $\mathbf{x}^{e_2}$ respectively. We can then generate synthetic OOD \(\tilde{\mathbf{x}} = D(\tilde{\mathbf{s}}, \mathbf{v}^{e'})\in\mathcal{D}_{OOD}\) that exhibit realistic variations (Figure~\ref{fig:oods}), making them valuable for training robust OOD detectors. Specifically, we employ an energy bounding mechanism to enhance the separability between ID and OOD instances, with the regularization term \( R_{E} \) defined as:
\begin{equation}
\small
\begin{aligned}
\label{eq:reg_ood}
    R_{E} 
    &= \mathbb{E}_{(\mathbf{x}_{i},y_i\in\mathcal{Y}_{tr})\sim\mathcal{D}_{tr}}\big( \max(0,Energy(\mathbf{x}_{i},\boldsymbol{\theta})-\gamma) \big)^2 \\
    &+ \mathbb{E}_{(\mathbf{x}_{j},OOD)\sim\mathcal{D}_{OOD}}\big( \max(0,\gamma-Energy(\mathbf{x}_{j},\boldsymbol{\theta})) \big)^2
\end{aligned}
\end{equation}

such that ID instances are encouraged to have energy smaller than \(\gamma < 0\), while OOD instances are encouraged to have energy larger than \(\gamma\).

\textbf{Loss Function.} The overall objective combines the feature and energy regularizations:
\begin{equation}
\small
\begin{aligned}
\label{eq:total_loss}
    \ell(\boldsymbol{\theta}) 
    = \mathbb{E}_{(\mathbf{x}^e, y^e) \sim \mathcal{D}_{tr}} \mathcal{L}_{CE}(f(\mathbf{x}^e, \boldsymbol{\theta}), y^e) + \zeta_1 \cdot R_{F} + \zeta_2 \cdot R_{E}
\end{aligned}
\end{equation}

where \(\mathcal{L}_{CE}\) represents the cross-entropy loss. 

\section{Experiments}

\begin{figure}[!t]
    \centering
    \includegraphics[width=0.8\linewidth]{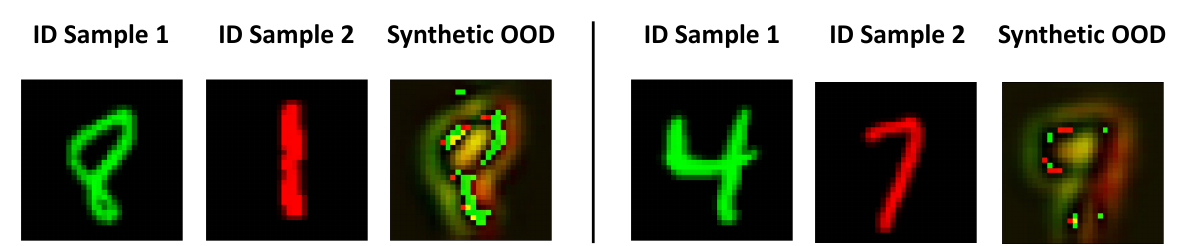}
    \vspace{-2mm}
    \caption{Examples of synthetic OODs generated by blending two ID samples.}
    \label{fig:oods}
    \vspace{-2mm}
\end{figure}

\begin{table}[!t]
\centering
\scriptsize
\caption{Performance Evaluation. The top scores are highlighted in \textbf{bold}, and the second-best scores are \underline{underlined}.}
\label{tab:summary-coloredmnist}
\begin{tabular}{l|c|c|c}
\toprule
Methods & \textbf{AUROC} & \textbf{AUPR} & \textbf{ID Accuracy} \\
\midrule
ERM* \cite{vapnik1999nature} & 52.27 $\pm$ \phantom{0}2.6 & 10.57 $\pm$ \phantom{0}0.4 & 35.39 $\pm$ 12.3 \\
IRM* \cite{arjovsky2019invariant} & 53.45 $\pm$ \phantom{0}0.2 & 11.35 $\pm$ \phantom{0}0.1 & 40.86 $\pm$ 14.6 \\
Mixup* \cite{yan2020improve} & 53.45 $\pm$ \phantom{0}1.4 & 10.95 $\pm$ \phantom{0}0.8 & 36.02 $\pm$ 13.0 \\
MBDG* \cite{robey2021model} & 63.01 $\pm$ \phantom{0}7.6 & 15.11 $\pm$ \phantom{0}2.6 & 52.15 $\pm$ 16.9 \\
\cmidrule{1-4}
EDir-MMD \cite{noguchi2023simple} & 47.16 $\pm$ \phantom{0}3.1 & 32.29 $\pm$ 22.5 & 60.99 $\pm$ \phantom{0}1.7 \\
EDst-MMD \cite{noguchi2023simple} & 53.47 $\pm$ \phantom{0}0.6 & 34.78 $\pm$ 23.7 & \underline{61.00} $\pm$ \phantom{0}2.8 \\
SCONE \cite{bai2023feed} & 59.23 $\pm$ \phantom{0}4.0 & 22.36 $\pm$ \phantom{0}1.3 & 55.60 $\pm$ 16.1 \\
DAML \cite{shu2021open} & 51.22 $\pm$ \phantom{0}0.4 & 35.05 $\pm$ \phantom{0}0.1 & 54.02 $\pm$ 21.6 \\
MEDIC \cite{wang2023generalizable} & 61.27 $\pm$ \phantom{0}4.6 & 36.69 $\pm$ \phantom{0}3.1 & 60.13 $\pm$ 15.3 \\
\cmidrule{1-4}
Ours-OCSVM & 72.15 $\pm$ \phantom{0}2.0 & 55.77 $\pm$ \phantom{0}0.3 & \textbf{71.03} $\pm$ \phantom{0}1.9 \\
Ours-DDU & 74.51 $\pm$ \phantom{0}2.4 & 42.29 $\pm$ \phantom{0}0.9 & \textbf{71.03} $\pm$ \phantom{0}1.9 \\
Ours-MSP & \underline{81.67} $\pm$ \phantom{0}2.5 & \textbf{57.39} $\pm$ \phantom{0}0.4 & \textbf{71.03} $\pm$ \phantom{0}1.9 \\
Ours-Energy & \textbf{81.92} $\pm$ \phantom{0}2.6 & \underline{57.23} $\pm$ \phantom{0}0.4 & \textbf{71.03} $\pm$ \phantom{0}1.9 \\
\bottomrule
\end{tabular}

\smallskip 
\raggedright 
\textit{* The best performance across all OOD detectors: OCSVM, DDU, MSP, and Energy.}
\end{table}

We evaluated our method on the \textsc{ColoredMNIST} \cite{arjovsky2019invariant} dataset following the setting of \cite{robey2021model}.
For each OOD digit selection, we conducted three trials, each including 30 random hyperparameter search runs. This process is repeated until at least 50\% of digits have been designated as OOD. 
For OOD regularization $R_{\text{OOD}}$, synthetic OODs are generated from ID data, particularly in low-density regions. Figure~\ref{fig:oods} shows examples of synthetic OODs where digit 9 is treated as the OOD class, highlighting that these synthetic OODs possess distinct semantic features compared to the ID data.
As shown in Table \ref{tab:summary-coloredmnist},
traditional DG methods (the first four) fail to learn robust domain-invariant features, while our methods consistently outperform all baselines, particularly in AUROC and AUPR. 
These results emphasize the effectiveness of our approach in enhancing both OOD detection and ID classification for open-set domain generalization.


\newpage
\bibliographystyle{bibtex/IEEEtran}
\bibliography{bibtex/IEEEabrv,bibtex/references}

\end{document}